\documentclass[a4paper,12pt]{article}
\usepackage{mathptmx}
\usepackage{amsmath,amstext,amsgen,amsbsy,amsopn,amsfonts,amssymb}
  \usepackage{setspace}
\usepackage{easybmat}
\usepackage{graphics}
\usepackage[pdftex]{graphicx}
\usepackage{textcomp}
\usepackage{gensymb}
\usepackage{epsfig}
\usepackage{amsthm}
\usepackage{bm}
\usepackage{algorithm}
\usepackage{algorithmic}

\usepackage{wrapfig}
\usepackage{esvect}
\usepackage{multirow}
\usepackage{array}
\usepackage{xcolor}
\usepackage{hyperref}

\usepackage[margin=2.88cm]{geometry}
\usepackage{txfonts}

\begin{document}

\large
\title{\textbf{Automatically Score Tissue Images Like a Pathologist by Transfer Learning}}
\normalsize
\author{
Iris Yan\dag
\vspace{0.1in}\\
\dag Amity Regional High School\vspace{0.06in}\\
Woodbridge, CT, USA
}
\date{}
\maketitle

\tableofcontents
\newpage

\begin{abstract}
Cancer is the second leading cause of death in the world. Diagnosing cancer early on can save many lives. 
Pathologists have to look at tissue microarray (TMA) images manually to identify tumors, which can be time-consuming,
inconsistent and subjective. Existing automatic algorithms either have not achieved the accuracy level of a pathologist 
or require substantial human involvements. A major challenge is that TMA images with different shapes, sizes, and locations can 
have the same score. Learning staining patterns in TMA images requires a huge number of images, which are severely 
limited due to privacy and regulation concerns in medical organizations. TMA images from different cancer types may share certain 
common characteristics, but combining them directly harms the accuracy due to heterogeneity in their staining patterns. 
Transfer learning is an emerging learning paradigm that allows borrowing strength from similar problems. However, existing
approaches typically require a large sample from similar learning problems, while TMA images of different cancer types 
are often available in {\it small sample size} and further existing algorithms are limited to transfer learning from one similar problem. 
We propose a new transfer learning algorithm that could learn from {\it multiple} related problems, 
where each problem has a small sample and can have a substantially different distribution from the original one. 
The proposed algorithm has made it possible to break the critical accuracy barrier (the 75\% accuracy level of pathologists), 
with a reported accuracy of 75.9\% on breast cancer TMA images from the Stanford Tissue Microarray Database. It is supported 
by recent developments in transfer learning theory and empirical evidence in clustering technology. This will 
allow pathologists to confidently adopt automatic algorithms in recognizing 
tumors consistently with a higher accuracy in real time.
\end{abstract}

\begin{keywords}
Tissue image scoring, Transfer learning, Small training sample, Multiple small auxiliary sets
\end{keywords}

\section{Introduction}
\label{section:introduction}
\begin{figure}[h]
\centering
\begin{center}
\hspace{0cm}
\includegraphics[scale=0.3,clip,angle=0]{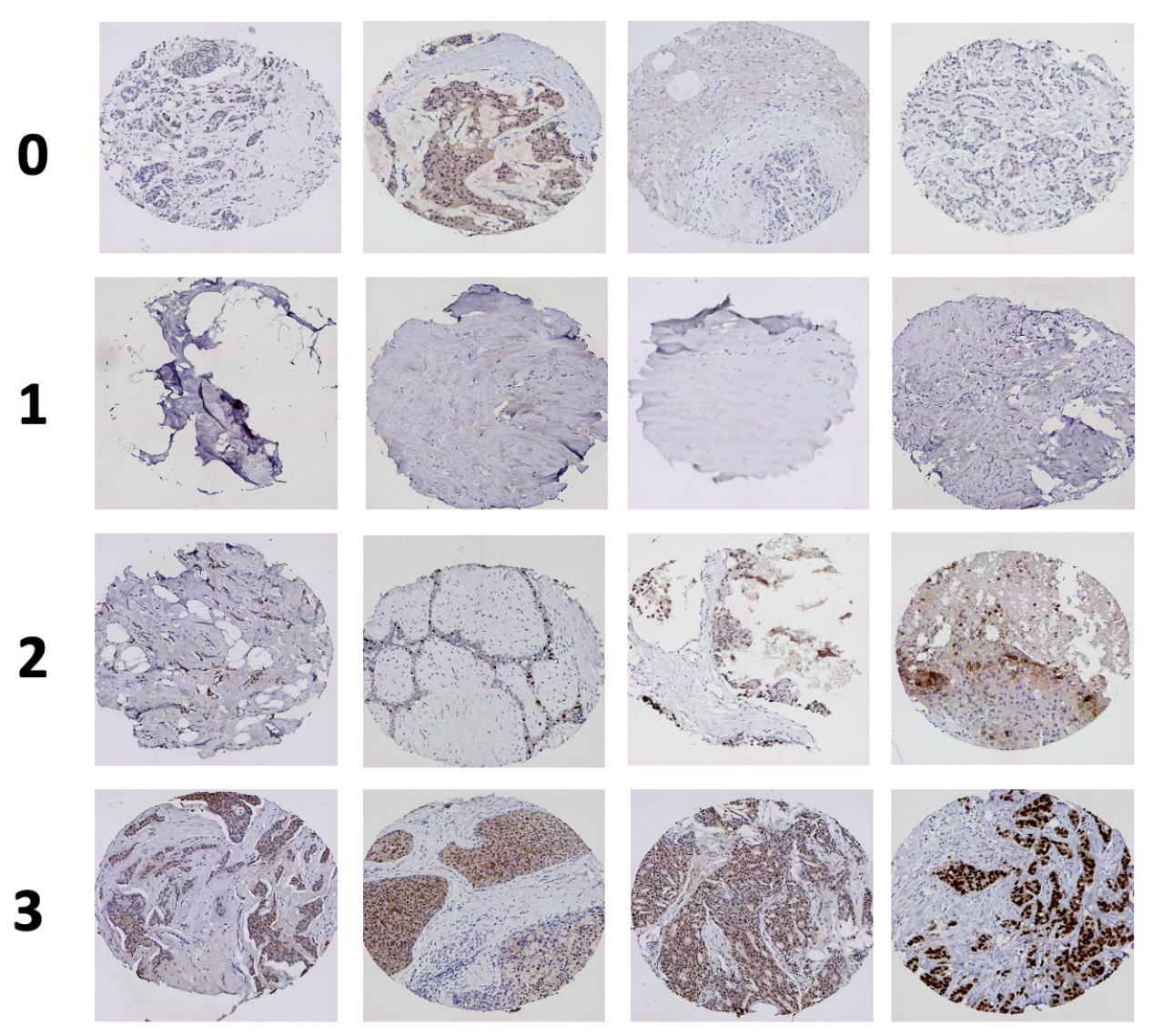}
\end{center}
\abovecaptionskip=-1pt
\caption{\it The staining patterns vary highly across TMA images. Images with the same score can look drastically different.} 
\label{figure:varTMA}
\end{figure}

Cancer continues to affect millions of people, yet many types of cancers,  e.g. pancreatic cancer, mesothelioma, gallbladder 
cancer, brain cancer etc, have very low survival rates. This has left treatments such as surgery, chemotherapy, 
and radiation therapy as the main options. As early cancer treatment significantly improves survival rates, the 
need for early diagnosis is imperative. Despite the importance and necessity of early diagnosis, it can take up to weeks 
before pathology reports are delivered. 
\\ 
\\
The tissue microarray (TMA) is an emerging technology to analyze tissue samples. It uses thin slices of tissue 
core samples that are arranged in an array format in paraffin blocks \cite{Grace-jones2012}. Biomarkers are applied and typically 
biomarker-specific dark colors of yellow stains are shown when there is presence of tumors. TMA images are then captured with a 
high-resolution microscope. The scoring of TMA images is based on the severity of tumors, and a common system for scoring 
TMA images uses a scale from 0 to 3, where a score of 0 indicates no tumors, and a score of 3 implies very severe tumors 
which corresponds to late stages of cancer progression, for example, Stage IV in a common 4-stage (Stage I, II, III, and IV) cancer grade system.
TMA technology makes it possible to efficiently analyze many tissue samples
together, thus normalizing conditions for comparative studies. These benefits give TMA images the potential to be widely used as 
an effective technique for diagnosis and prognosis oncology \cite{Jawhar2009, PageMagliocco2014}. A valuable 
source to explore TMA images is the Stanford TMA image database \cite{Marinelli2007}, publicly available 
from \url{https://tma.im/tma_portal/}, and work presented here is based on TMA images from this database.
\\
\\  
The diagnosis for tumors---identifying tumors from staining patterns in tissue images---manually is time-consuming and can easily 
be inconsistent \cite{CampNR2008,Vrolijk2003}. Although structured procedures have been proposed, manual scoring can still be 
fairly subjective \cite{TACOMA, BrunyeMercan2017}, and the same pathologist may score differently at different sessions for the 
same image.  Because of these issues, a number of algorithms, including ACIS, Ariol, TMALab, AQUA \cite{CampRimm2002} and TACOMA \cite{TACOMA}, 
have been developed to automate this process. These algorithms, although transformative, are not widely adopted 
due to limits in their capabilities. They require substantial involvement of pathologists on tasks such as TMA image background subtraction, 
feature segmentation, thresholds of hue or pixel intensity, and the provision of representative TMA image patches etc.  
\\ 
\\
The variability of staining patterns in TMA images and the scarcity of training samples make it particularly challenging to develop
an automatic algorithm. Staining patterns in TMA images can have very different sizes, locations, shapes, and colors, despite having 
the same score. Figure~\ref{figure:varTMA} demonstrates the variability of staining patterns in the TMA images. 
The high variability in the staining patterns also increases the need for larger sample sizes in training algorithms. Having more 
images would allow the algorithm 
to capture more variability in the staining patterns, leading to more consistent results. However, the number of TMA 
images is severely limited due to privacy and regularization concerns in medical organizations. Moreover, the TMA images 
currently available are for over 100 different types of cancers, with very few numbers of TMA images available for each individual cancer type.
In recent years, deep learning \cite{GoodfellowBengioC2016} has become the method of choice for many image recognition tasks. 
However, that is not feasible for TMA image scoring due to the small sample size available for a given cancer type.  
\\
\\
This work aims to develop an automatic algorithm for the scoring of TMA images at the accuracy level of pathologists by augmenting the training 
sample with transfer learning \cite{Caruana1997}. 
The key observation is that, some of the TMA images of a different cancer type may look similar (i.e., have a similar staining pattern) to 
those of the given cancer type and with the same score, thus making it possible for transfer learning.  However, usual algorithms for transfer 
learning (many based on deep neural networks \cite{TanSunKong2018}) are not applicable since we do not have a large dataset from
similar problems as the basis for knowledge transfer---while there exist many different cancer types, the number of TMA images of each 
individual type is small. This gives rise to a {\it new} transfer learning setting: there are multiple similar learning problems
around but each has only a small training sample, and we wish to design an algorithm that could effectively transfer knowledge
from all the similar learning problems. The approach we take is instance-based transfer learning \cite{Caruana1997}, where we propose 
to selectively include TMA images of other cancer types with similar staining patterns as the cancer type of interest. With an enlarged training set, we
can expect to improve accuracy on the scoring of TMA images of the original cancer type. Given the huge success of transfer learning in 
application domains such as natural language processing and image recognition, we expect that it would enable the automatic 
evaluation of cancer tumors in TMA images at the level of pathologists. 
\\
\\
The remaining of this paper is organized as follows. In Section~\ref{section:method}, we describe our approach and algorithms design. 
Experiments and results are presented in Section~\ref{section:results}, along with a discussion of connections of our algorithm to recent
theoretical developments in transfer learning. Finally we conclude in Section~\ref{section:conclusions}. 
%
\section{Methods}
\label{section:method}
We formulate the scoring of TMA images as a classification problem.
Our approach can be summarized as follows. A type of tumor-specific spatial histograms is used to capture key features 
characterizing the staining patterns in tissue images, then transfer learning is adopted to increase the training sample 
size by selectively including images of other cancer types. The enlarged training set is input to Random Forests (RF) \cite{RF} 
classifier to re-fit the classification model and results reported. The goal is to achieve or exceed the accuracy level of pathologists. 
A detailed description about spatial histogram, RF, transfer learning, and the algorithmic implementation of our approach 
will be presented in Sections \ref{section:spatialHistogram}, \ref{section:vignetteRF}, \ref{section:transferLearning}, and \ref{section:algorithmDesc}, respectively. 
\begin{figure}[h]
\centering
\begin{center}
\hspace{0cm}
\includegraphics[scale=0.33,clip,angle=0]{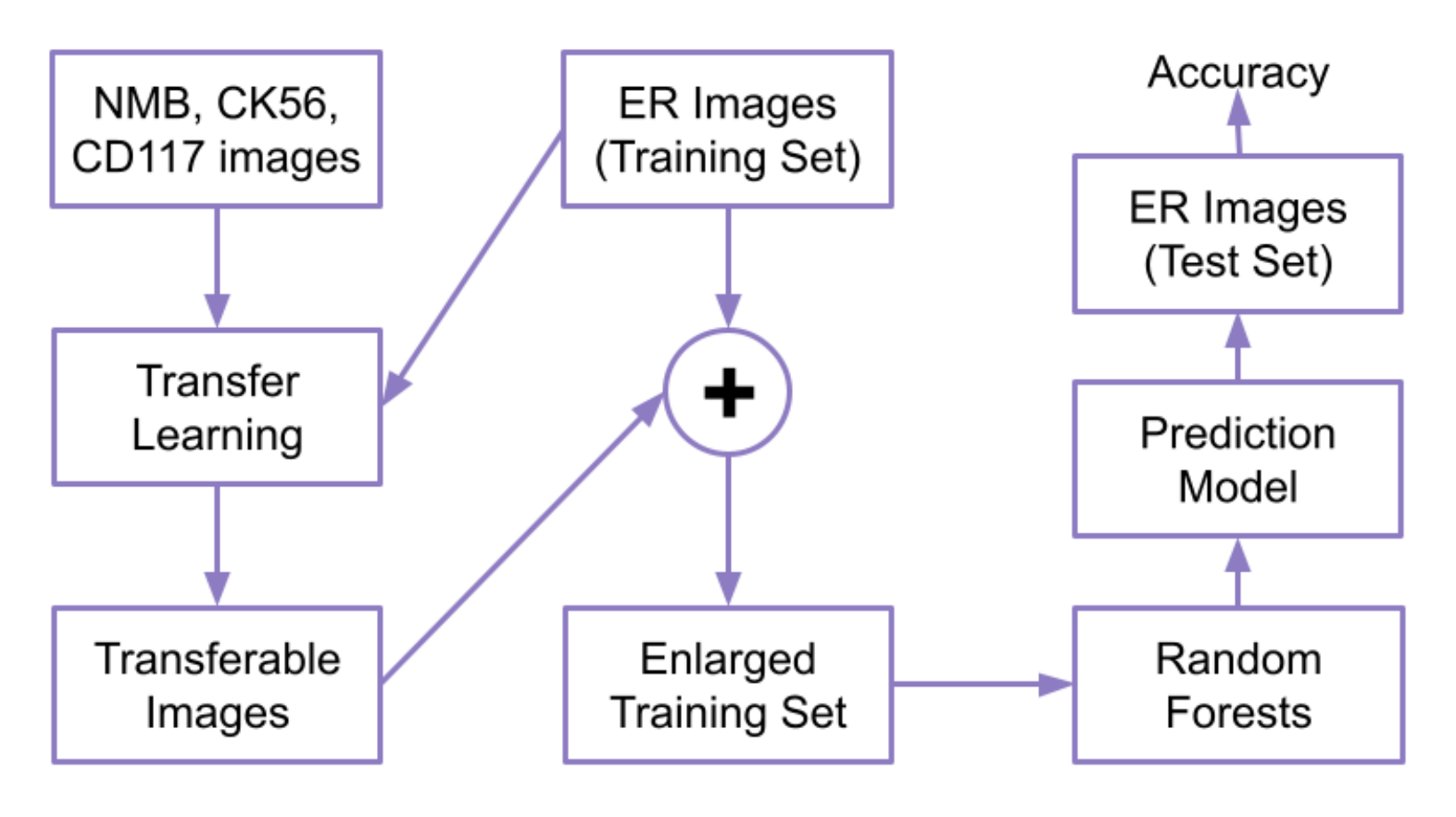}
\end{center}
\abovecaptionskip=-1pt
\caption{\it  Illustration of the overall flow of the proposed algorithm. The '+' sign stands for combining images from multiple sources. 
ER is the name of biomarker associated with the target cancer type and we use it to indicate the corresponding TMA images, 
while NMB, CK56 and CD117 are those for other cancer types. } 
\label{figure:projFlow}
\end{figure}
\\
Figure~\ref{figure:projFlow} shows the overall flow of our approach. Note that here for illustration 
purpose, we use `ER', the name of a biomarker for breast cancer, to indicate the cancer type of interest, while using 'NMB', `CK56', 
or `CD117' for other cancer types. One major challenge in the automatic scoring of TMA images is the variability in the staining patterns, which may have different shapes, sizes, and locations despite having the same score. The staining patterns are encoded by spatial histograms, following work in \cite{TACOMA}. 
The spatial histogram also helps to greatly reduce 
the data dimensionality, another challenge in the automatic scoring of TMA images which is caused by
large image sizes (i.e., $1504 \times 1410$). 
RF is used as the classifier for its strong built-in feature selection capability. 
To implement transfer learning, our main effort lies in 
evaluating whether an image of other cancer types is conformal to the hypothesis (or decision rule) induced by images of the
original cancer type. 

\subsection{The spatial histogram}
\label{section:spatialHistogram}
The staining patterns on TMA images can vary highly, even for those with the same score. Thus, it is desirable to look for image 
features that are relatively stable across images of the same score. The feature we use is a spatial histogram matrix (or gray level 
cooccurrence matrix in the remote sensing literature \cite{Haralick1979}), which is commonly used for textured images. It is a 
suitable image feature because TMA images belong to one of the two major classes of textured images---images taken from very far away 
in the sky (for example, remote sensing images) and those taken under a high resolution microscope. 
Indeed, the spatial histogram has been used in literature for TMA images \cite{TACOMA} and fares well. 
\\ 
\\
Similar to the conventional histogram, the spatial histogram is a collection of counting statistics about pairs of adjacent 
(or neighboring) pixels in an image and represented as a matrix. 
\begin{figure}[h]
\centering
\begin{center}
\hspace{0cm}
\includegraphics[scale=0.1,clip,angle=0]{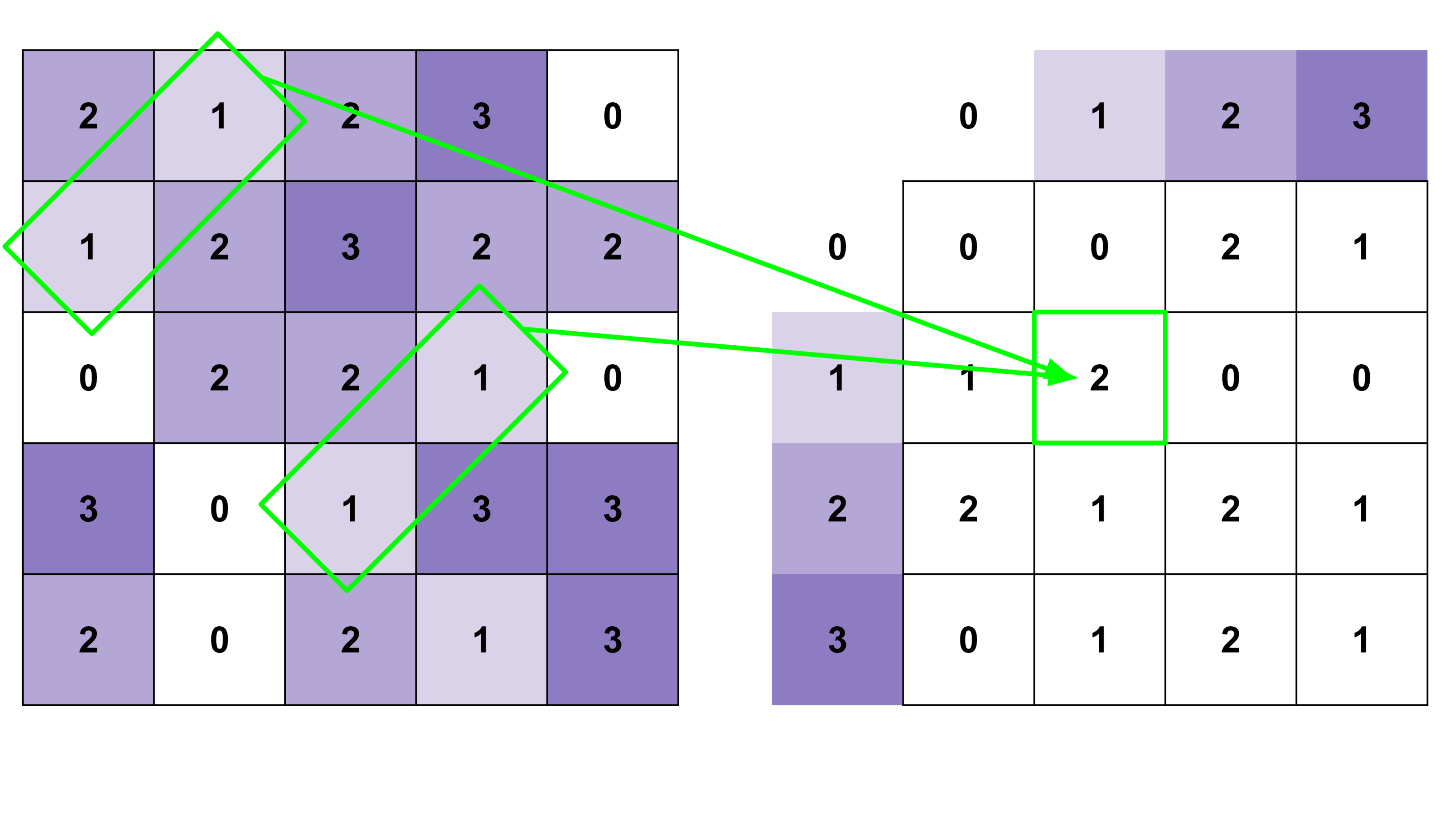}
\end{center}
\abovecaptionskip=-6pt
\caption{\it Illustration of a toy image and the spatial histogram matrix. The left panel stands for the original image where 
the numbers are the gray values, and the right is the resulting spatial histogram matrix with a dimension $4 \times 4$. 
The two diagonally (i.e., along 45 \degree direction) neighboring pixels with both gray levels of 1 occur twice in the image, 
so the (1, 1)-entry of the spatial histogram matrix has a value of 2. } 
\label{figure:glcm}
\end{figure}
In generating a spatial histogram, we only consider neighboring pixels that have a predefined spatial relationship. That is, 
the two pixels need to have a pair of given pixel values and are neighbors along a certain direction. A typical choice for the 
direction is the $45\degree$ direction in the image grid. 
Figure~\ref{figure:glcm} shows an example of such neighboring pixels where the pairs of pixels of interest are marked by 
green rectangles and both have grey value 1. The counting statistic is the total number of times such a pair 
occur in the image grid along the $45\degree$ direction, and this gives the value of the (1,1)-entry of the spatial histogram matrix. 
Similarly, pairs of pixels along the same direction but with pixel values $i$ and $j$, respectively, would produce a counting statistic for 
the (i,j)-entry of the matrix, and so on.
\\
\\
One can choose to use a different spatial relationship, like pairs along a direction of $0\degree$ or $135\degree$ etc in the image
grid, and the difference in TMA image scoring would be small. Also, one can extend the distance between the neighboring pixels.
The default distance is 1, indicating immediate neighbors. The two pairs of example pixels in Figure~\ref{figure:glcm} 
all have a distance 1, while a distance of 2 or larger would indicate a larger staining pattern. In this work, a spatial distance of 2
is used. 
\\
\\ 
One nice feature about the use of spatial histogram is dimension reduction. The TMA images are large in size, and those 
from the Stanford TMA image database \cite{Marinelli2007} have a size of $1504 \times 1440$. Directly working with such images
would require enormous computing power and memory, and worse still, that will lead to the curse of dimensionality, as 
each image would be treated as a huge vector of a dimension more than 2 million (i.e., $1504 \times 1440$). As the gray value of
TMA image pixels have a range between 0 and 255, the spatial histogram will reduce the data dimension to a value, $256 \times 256$,
much smaller than that from the original image. One step further is to apply a quantization to the image gray values. That will lead to an even 
smaller spatial histogram matrix. We follow work in \cite{TACOMA}, and apply a linear quantization to the gray values into 51 levels. That is,
a gray value of $5\cdot x+y$ for $0\leq x\leq50$ and $0\leq y \le 4$ will be transformed to $x+1$ (255 is converted to 51 for simplicity). 
Thus the spatial histogram matrix now has a dimension of $51 \times 51$,  and a vector (of dimension 2601) formed by 
collapsing this matrix is used as input to the RF classifier so that computation can be done very efficiently. 
Importantly, 
this also makes the resulting image features stable against small variations in image pixel values due to varying physical conditions 
such as lighting or illumination. 
\subsection{Random Forests}
\label{section:vignetteRF}
RF is used as the classification engine for our TMA image scoring algorithm.
RF is an ensemble of decision trees, and each tree is grown by recursively splitting 
the data. At each node split, RF randomly samples a number of features (called number of tries) and selects 
one leading to their best partition of that node, according to some criterion. Each tree progressively narrows down
the decision for an instance. The node split continues until there is only one point in the node  (for classification) or 
when the node is pure (i.e., all the points in the node have the same label). At classification, an instance receives 
a vote on its class label from each tree in RF, and the final decision given by RF is a majority vote on the class labels,  according 
to the number of votes that each label gets from all trees.
\\
\\
Many studies have reported excellent performances of RF
\cite{RF,caruanaKY2008}. For TMA image classification, previous studies \cite{HolmesKapelner2009, TACOMA} also 
show that RF outperforms competitors like SVM \cite{SVM} or boosting \cite{FreundSchapire1996}. 
Compared to its competitors, RF scales well against some main challenges in TMA image scoring, including high dimensionality
and label noise, thanks to its strong feature selection ability and ensemble nature.
RF is easy to use with very few tuning parameters---often one just need to set the number of trees and the number 
of tries at each node split.
\subsection{Transfer learning}
\label{section:transferLearning}
\begin{figure}[t]
\begin{center}
\hspace*{-0.1cm}
\includegraphics[scale=0.32]{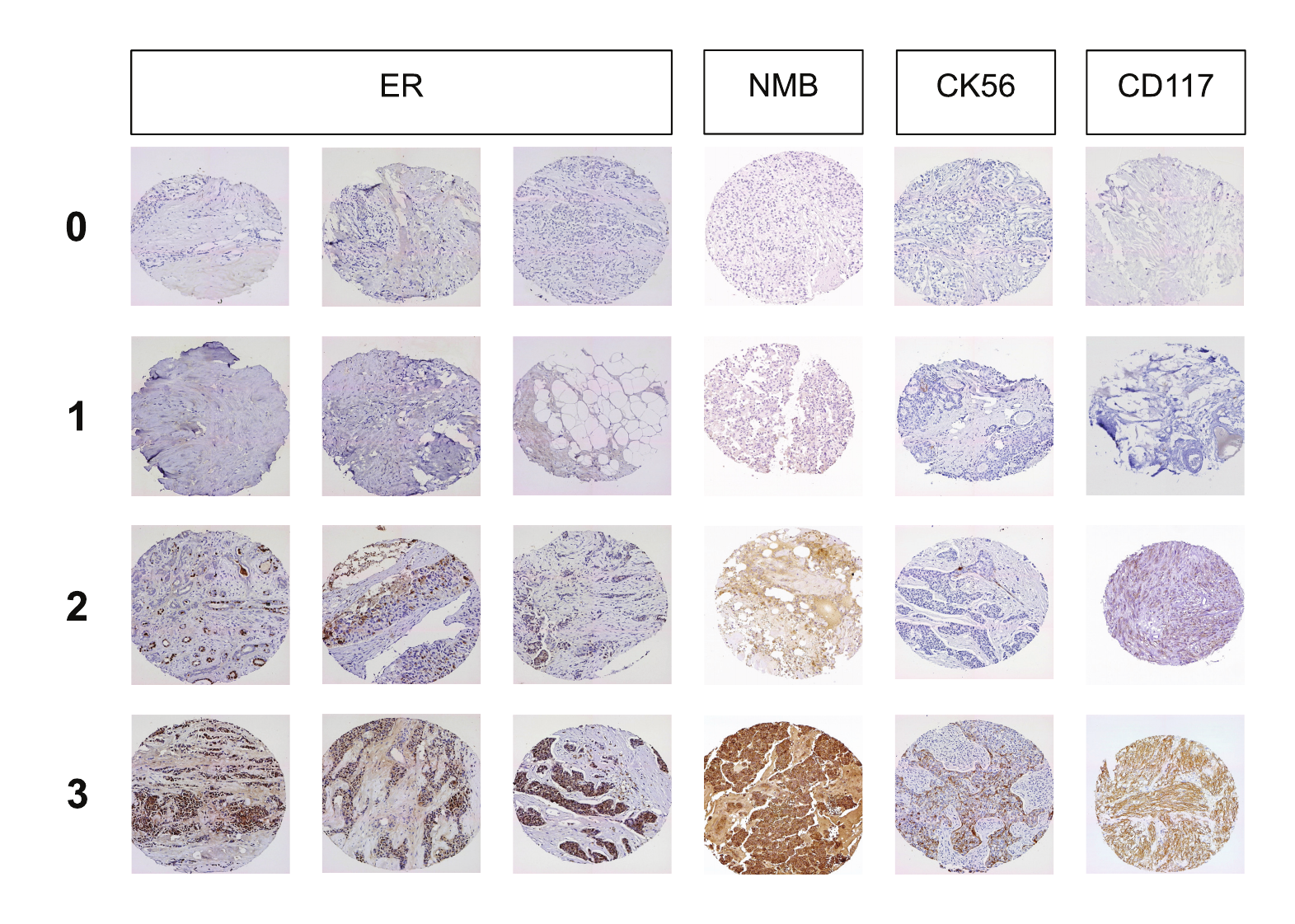}
\end{center}
\abovecaptionskip=-1pt
\caption{\it Example transferable images from other cancer types. The left 3 columns of images are TMA images for breast cancer and indicated by the 
associated biomarker estrigen receptor (ER). The right 3 columns are TMA images for cancer types indicated by biomarkers NMB, CK56 and 
CD117, respectively, which have a similar appearance as those for ER with the same label. } 
\label{figure:cancertypesTL}
\end{figure}

Transfer learning \cite{Caruana1997} is an emerging learning paradigm to address the problem of insufficient training 
data when there is a large set of auxiliary data (called {\it auxiliary set}) that entails 
knowledge helpful in solving the original problem. Transfer learning algorithms can be classified as instanced-based, mapping-based, 
representation-based, or feature-based transfer learning \cite{TanSunKong2018}. Instance-based transfer learning \cite{DaiYang2007} 
transfers knowledge in the form of enlarging the original training set by finding instances in the auxiliary set that are 
consistent with the hypothesis learned on the original training set (such instances are called {\it transferable}). 
Mapping-based transfer learning \cite{TzengDarrel2014} learns semantically sensible invariant representation across 
the original and auxiliary sets. Feature-based approaches \cite{LongJordan2015} learn features that would help the 
learning of the original problem. Representation-based \cite{GAN2014, OquabBottou2014} tries to find representations 
that can be transferred. Recently, deep transfer learning \cite{GoodfellowBengioC2016} has become very popular and 
achieves impressive performance in a number of domains, for example, large natural language processing systems 
such as BERT \cite{DevlinLee2019} and GPT-3 \cite{GPT3-2020}, and pre-trained image models \cite{ShafahiStuder2020}. 
The literature on transfer learning is enormous, and we can only mention a few here. More discussion can be seen in 
\cite{PanYang2010, TanSunKong2018, AgarwalSondhi2021} and references therein.
\\
\\
The lack of a large auxiliary dataset makes transfer learning particularly challenging for the problem of TMA image scoring. 
The big family of deep neural networks based transfer learning algorithms are not applicable here due to the reliance on 
the training of large deep networks, which inevitably requires a huge training set. In the scoring of TMA images, there are 
multiple auxiliary sets available as images from a number of other cancer types can look very similar to those of the
cancer type of interest. However, none of the auxiliary sets is large enough
for the typical deep neural networks based approaches to be feasible. Thus, we now have a {\it new problem setting for transfer
learning}, and we wish to enable knowledge transfer from {\it multiple small} auxiliary sets. 
\begin{figure}[h]
\begin{center}
\includegraphics[scale=0.28,clip,angle=0]{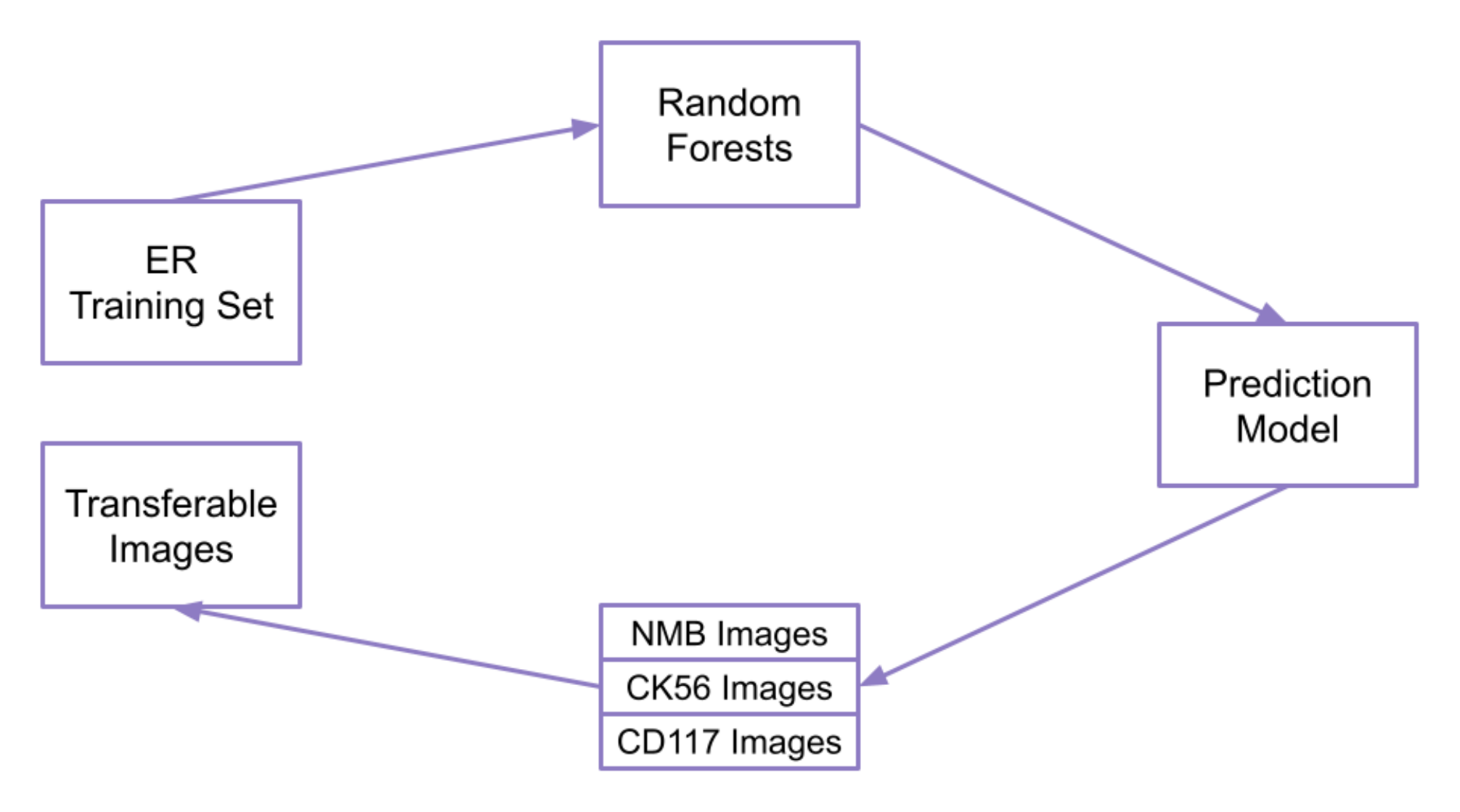}
\end{center}
\abovecaptionskip=-1pt
\caption{\it Illustration of transfer learning. ER is the name of biomarker associated with the target cancer type, 
while NMB, CK56 and CD117 are those for other cancer types. }
\label{figure:transferLearning}
\end{figure}
\\
\\
The approach we take is instance-based transfer learning, and Figure~\ref{figure:transferLearning} is an illustration of the algorithm. 
We first fit a prediction model (called original hypothesis) using RF on the original training set. Then from the auxiliary 
set, we try to identify TMA images that are consistent with the original hypothesis. Clearly we do not require a large 
auxiliary set to achieve this. We can add 
those transferable images to enlarge the original training set. For a small 
training set, increasing its size will likely improve performance on the test set. In Figure~\ref{figure:cancertypesTL}, the left 3 columns
show example images for the target cancer type---breast cancer (indicated by ER) where each row corresponds to a different score. The right
columns are example images from a different cancer type (marked by NMB, CK56 and CD117, respectively) that look similar (to certain extent) 
to the breast cancer images with the same score. 
\subsection{An algorithmic description}
\label{section:algorithmDesc}
In our algorithm, transfer learning is implemented as function, $tmaTransfer()$, which finds transferable images from 
a given auxiliary set. The main function, $tmaScore()$, calls $tmaTransfer()$ to obtain transferable images from other
cancer types, fits model on the enlarged training set via RF, and then reports final results on the test set. 
\\
\\
$tmaTransfer()$ is implemented as follows. We first fit a classification model, $\mathbb{M}_0$, by RF on the original 
training set $\mathcal{T}_0$. Then we apply $\mathbb{M}_0$ to the set of auxiliary images, $\mathcal{W}$. If the
predicted label for an image, $W \in \mathcal{W}$, is the same as its given label (note the label for these images in the 
auxiliary set are known since they come with the TMA images in the Stanford database), then we say image $W$ (along with 
its label) is consistent with the original model hypothesis $\mathbb{M}_0$. As there might be label noise for image $W$, we only 
transfer images that are predicted more confidently. The confidence in prediction can be estimated from the number 
of votes $W$ receives on different class labels (or scores) from trees in RF. If the majority class gets substantially more votes 
than other classes, then we say the instance is predicted with high confidence. Here, the majority class is the one
that receives the most votes. An easy way to estimate the confidence is to use the difference in the fraction of votes received
for the top and the second majority class (the class that receives the second most number of votes). Let $n_1(W)$ and $n_2(W)$ 
be number of votes $W$ gets on the top and second majority class, and $T$ be the number of trees in RF. We can estimate 
the confidence in predicting instance $W$ as follows
\begin{equation}
\beta(W) = \frac{n_1(W) -n_2(W)}{T}.
\end{equation}
So $\beta(W)$ has a value in the range $[0,1]$, and a TMA image $W$ in the auxiliary set is selected if it is classified with a 
confidence larger than a predetermined level $\beta_0$. The choice of $\beta_0$ seeks to include TMA image instances that 
are valuable to the original problem. Singh and his coauthors \cite{SinghNowarkZhu2009} study the contribution of individual 
data points to algorithmic performance in semi-supervised classification problem, and they find that data points that are along 
the decision boundary barely help, while the value of a data point increases when the data point is slightly away from the boundary. 
Our definition of confidence aims to avoid data points that are along or very 
close to the decision boundary (such data points would be classified with very low confidence), while trying to include data 
points that are slightly away from the decision boundary. Note that a too big value of $\beta_0$ is not desirable either, as that 
will cause the inclusion of only data points far away from the decision boundary. Let $\mathcal{F}$ denote the transferable 
set (which is the set of images transferred from the auxiliary sets). The $tmaTransfer()$ 
function is implemented as Algorithm~\ref{algorithm:tmaTransfer}.
\begin{algorithm}
\caption{\it~~tmaTransfer($\mathbb{M}_0, \mathcal{W}, T, \beta_0$)}
\label{algorithm:tmaTransfer}
\begin{algorithmic}[1]
\STATE Initialize the transferable set $\mathcal{F} \gets \emptyset$; 
\WHILE {$\mathcal{W}$ is not empty}
\STATE Pick an image $W$ from $\mathcal{W}$, and set $\mathcal{W} \leftarrow \mathcal{W} \setminus \{W\}$;
\STATE Apply the original model $\mathbb{M}_0$ to $W$;
\IF{predicted label on $W$ is different from its given label} 
	\STATE Skip to the next round of the loop; 
\ENDIF 
\STATE Calculate the prediction confidence $\beta(W)$ for image $W$;
\IF{$\beta(W)  \geq \beta_0$} 
	\STATE Add $W$ to the transferable set, $\mathcal{F} \leftarrow \mathcal{F} \cup \{W\}$; 
\ENDIF 
\ENDWHILE
\STATE return($\mathcal{F}$); 
\end{algorithmic}
\end{algorithm} 
\\
\\
To describe algorithm for the main function $tmaScore()$, assume that the other cancer types for transfer learning 
are associated with biomarkers NMB, CK56 and CD117 for simplicity of description. Let the set of auxiliary images for 
these cancer types be 
denoted by $\mathcal{W}_{nmb}, \mathcal{W}_{ck56}, \mathcal{W}_{cd117}$, respectively. Function $tmaScore()$ first fits a 
prediction model, $\mathbb{M}_0$, from the original training set using RF. Then, it identifies the transferable set from each
auxiliary set in $\{\mathcal{W}_{nmb}, \mathcal{W}_{ck56}, \mathcal{W}_{cd117}\}$. Add the transferable set to 
the original training set $\mathcal{T}_0$, re-fit the prediction model, then apply it to the test set and report results. The
main function is implemented as Algorithm~\ref{algorithm:tmaScore}.
\begin{algorithm}
\caption{\it~~tmaScore()}
\label{algorithm:tmaScore}
\begin{algorithmic}[1]
\STATE Let the number of trees in Random Forests be $T$;
\STATE Apply RF to the original training set $\mathcal{T}_0$; 
\STATE Let the fitted Random Forests model be $\mathbb{M}_0$;
\STATE Pick a predefined confidence level $\beta_0$;
\STATE Initialize the transferable set $\mathcal{F} \gets \emptyset$; 
\FOR {$\mathcal{W}$ in $\{\mathcal{W}_{nmb}, \mathcal{W}_{ck56}, \mathcal{W}_{cd117}\}$}
	\STATE Apply transfer learning to image set $\mathcal{W}$;
	\STATE $\mathcal{F}_t \leftarrow tmaTransfer(\mathbb{M}_0, \mathcal{W}, T, \beta_0)$;
	\STATE Add $\mathcal{F}_t$ to the transfer set $\mathcal{F} \leftarrow \mathcal{F} \cup \mathcal{F}_t$;
\ENDFOR
\STATE Add transferable set $\mathcal{F}$ to original training set $\mathcal{T}_0$ and re-fit RF;
\STATE Apply the re-fitted model to test set $T_{s}$ and report accuracy;  
\end{algorithmic}
\end{algorithm} 
\section{Experiments and results}
\label{section:results}
We conduct experiments using TMA images from the Stanford TMA image database. The cancer type we choose 
to work with is breast cancer, due to the fact 
that it is one of the best understood cancer types to date. The associated biomarker is estrogen receptor (ER). 
There are 690 images in total for ER in the database, and the training and testing sets are randomly split evenly. The reported 
results are averaged over 100 runs. 
\\
\\
TMA images in the Stanford TMA database are from several dozens of different cancer types. One can use 
TMA images from all other different cancer types, but we take a more conservative approach. Transfer learning 
requires the auxiliary set to be consistent to hypothesis entailed by the original training set. We browse through 
the Stanford TMA image database, and determine that TMA images associated with biomarkers NMB, CK56, and CD117 
are visually, to a certain extent, similar to TMA images for ER. Note that the actual set of images 
that can be used as auxiliary set, or images that are transferable, are broader that those visually similar images.
In other words, some of the transferable images may not look so ``similar" to images in the original training set. 
Part of our goal in using the term `visually similar images' is 
for the purpose of describing our motivation with transfer learning.
\\
\\
We use the R package {\it randomForest} for our experiments. The number of tries at each node split 
is chosen among $\{\sqrt{p}, 2\sqrt{p}\}$, where $p$ is the input data dimension. For TMA images as we work on the 
spatial histogram matrix, we have $p=51\cdot51=2601$. The number of trees in RF is fixed at $T=100$,  
adopting the value used in \cite{TACOMA}; indeed we find the difference in results small when varying over a range of 
choices $\{50, 100, 200, 500\}$. The confidence level $\beta_0$ for instance transfer is picked as 10\%, implying that only instances
with the top majority class leading the second majority class by at least 10\% votes (out of $T=100$ trees) are considered to be transferable. 
\subsection{Results}
\begin{figure}[h]
\centering
\begin{center}
\hspace{0cm}
\includegraphics[scale=0.47,clip,angle=0]{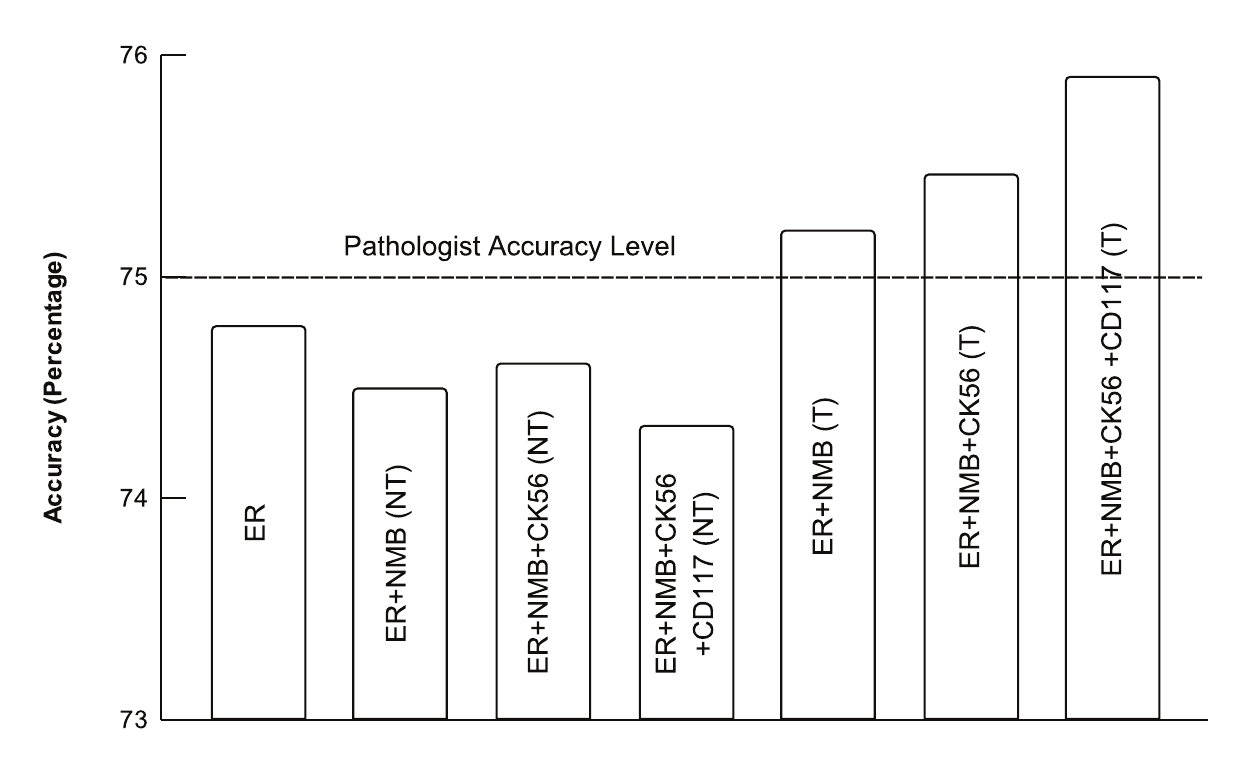}
\end{center}
\abovecaptionskip=-1pt
\caption{\it Comparison of accuracy. 'T' and 'NT' stand for transfer learning 
and without transfer learning, respectively, and ER, NMB, CK56, and CD117 indicate
the TMA images associated with respective biomarkers.} 
\label{figure:results}
\end{figure}

The evaluation metric is the test set accuracy, which is the percent of test images with a predicted class label agreeing 
with the given one (that is, the label comes with the database). The results are shown in Figure~\ref{figure:results}. 
The accuracy achieved with transfer learning over auxiliary sets associated with NMB, CK56 and CD117 is 75.9\%, 
outperforming the algorithm without transfer learning (shown as the first bar in the figure). The accuracy 
of pathologists is estimated to be around 75\--84\%  \cite{TACOMA}, so transfer learning allows our algorithm to achieve 
the accuracy level of pathologists. It should be noted that the gold standard of comparison vs pathologists in this 
case would be through consensus scores 
produced by a group of well-trained pathologists. However, in the lack of consensus pathologists' scores, 
estimation by \cite{TACOMA} could be viewed as giving the {\it accuracy level} of pathologists (on the ER images).
It is interesting to see that the achieved accuracy increases progressively when we apply transfer 
learning over more auxiliary sets, e.g., in the order of over $\mathcal{W}_{nmb}$ only, over two auxiliary sets $\{\mathcal{W}_{nmb}, \mathcal{W}_{ck56}\}$, 
and over three auxiliary sets $\{\mathcal{W}_{nmb}, \mathcal{W}_{ck56}, \mathcal{W}_{cd117}\}$. Note that comparison with other popular classifiers such as support vector machines and boosting with the spatial histogram matrices 
as inputs was made previously in \cite{TACOMA} for which an accuracy at 65.24\% and 61.28\% were reported, respectively. The reason why SVM and boosting substantially underperformed is possibly due to the fact that these two classifiers are not able to handle high dimensional and possibly noisy (in labels) inputs well, while RF has strong built-in ability in feature selection with high dimensional inputs and is also remarkably resistant to label noises.
\\
\\
We also conduct experiments by simply combining the training set of TMA images for ER with images in auxiliary sets 
$\mathcal{T}_{NMB}$, $\mathcal{T}_{CK56}$ and $\mathcal{T}_{CD117}$. This actually leads to a decrease in 
accuracy compared to that without transfer learning, as shown in the second thru the fourth bars in Figure~\ref{figure:results}. 
Although directly combining data from the auxiliary sets greatly increases the size of the training set, it also makes the data a lot more heterogeneous 
thus more challenging for classification as we now have to accommodate images of sub-models within the 
same class label. In comparison, transfer learning with our approach over images from other cancer types, even with different distributions, 
allows to improve the accuracy if we can properly control the confidence level. 
\subsection{Understanding the transfer learning scheme}
\begin{figure}[h]
\centering
\begin{center}
\hspace{0cm}
\includegraphics[scale=0.4,clip,angle=-180]{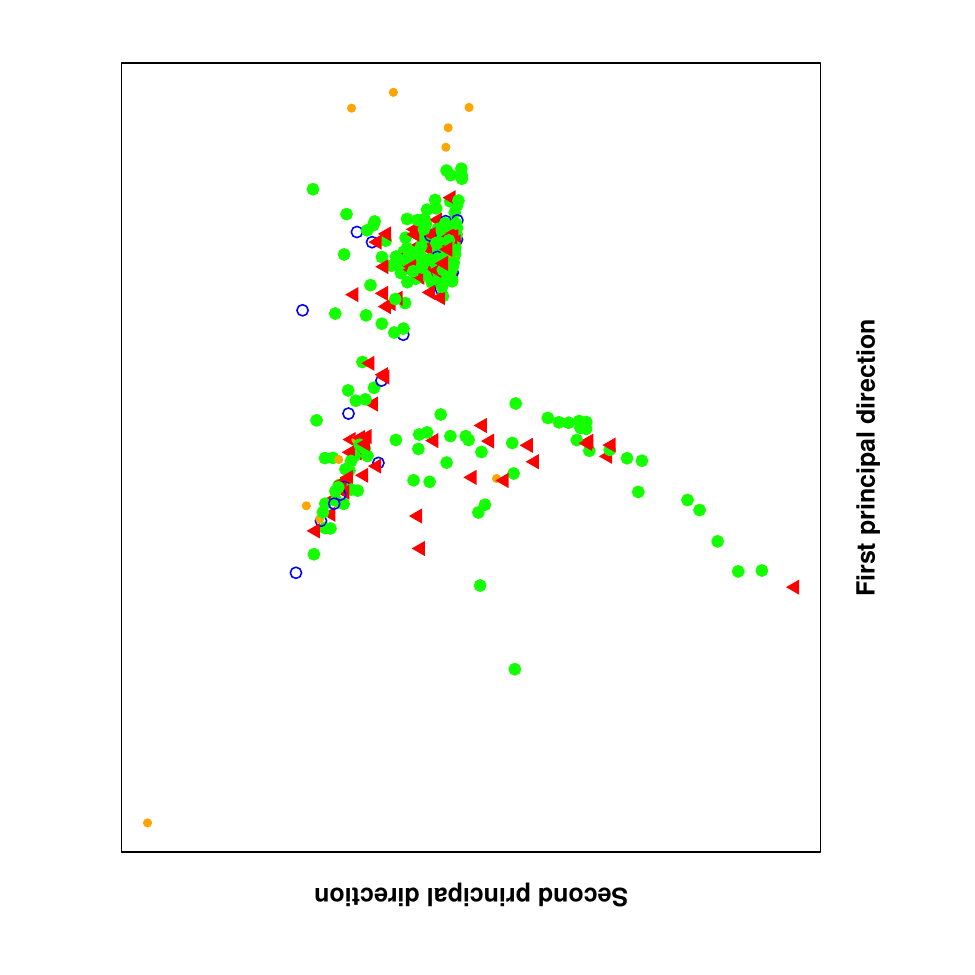}
\includegraphics[scale=0.4,clip,angle=-180]{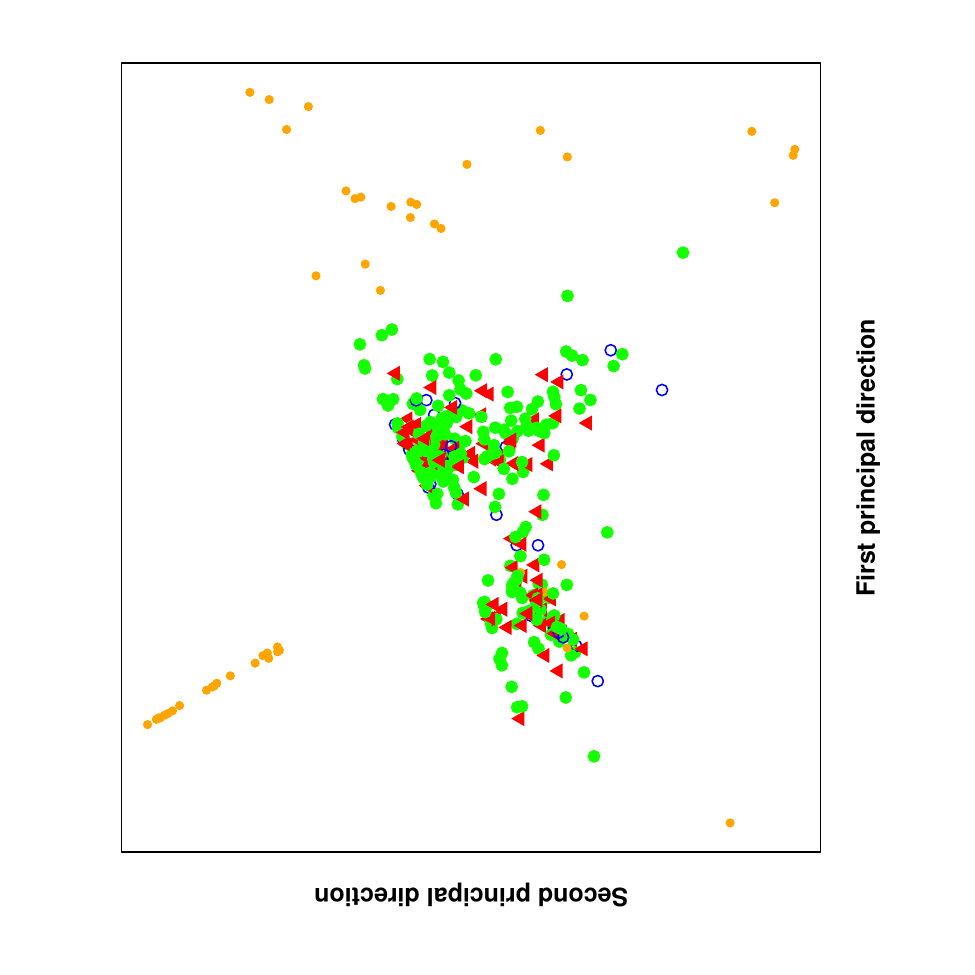}
\end{center}
\abovecaptionskip=-1pt
\caption{\it The original and transfer learning enlarged training set visualized by their first and second principal directions via principal
component analysis. The top panel is for the original training set, and the bottom panel is for training set enhanced by transfer
learning. Different colors correspond to TMA images with a different class label (or score). } 
\label{figure:pcaVisual}
\end{figure}

We adopt instance-based transfer learning, and the particular scheme we propose improves the accuracy of TMA image classification. 
Our algorithm could be understood from recent theoretical developments in transfer learning, and is also empirically supported 
by experiments. 
\\
\\
Transfer learning typically requires similarity in the distribution between the original and the auxiliary set. 
Recent work towards understanding of transfer learning focuses mainly on relaxing this ideal condition along two lines. 
One is the covariate-shift model where the marginal distribution of the original and the auxiliary data are different 
but their induced decision rules (or hypothesis) are similar. Here the marginal distribution refers to the probability distribution of the 
TMA images or their spatial histograms, while the induced decision rule is the rule that decides which class label (score) 
a TMA image gets given the pixel values or the spatial histogram of an image. Kpotufe and Martinet \cite{KpotufeMartinet2021} 
study, under the covariate-shift model, how much the target performance is impacted by sample size and the difference 
in the original and the auxiliary distribution.
The other is the posterior-drift model where the marginal distribution of the original and the auxiliary 
data are similar but their induced decision rules could be very different. Cai and Wei \cite{CaiWei2021} study
how fast the estimated decision rule converges to its limit in terms of the difference in the induced 
decision rules between the target and auxiliary data. For the scoring of TMA images, clearly the distribution of the original 
and the auxiliary data are different, and so are the induced decision rules. Our approach can be viewed as trying to 
satisfy the assumption of the covariate-shift model, i.e., it tries to find a subset of the auxiliary data such that the 
induced decision rule agrees to that from the original data. This is achieved by searching from the auxiliary set
those TMA images with the same label as the predicted one under the decision rule learned from the original data
(i.e., conformal to the original hypothesis).
This effectively overcomes the difficulty in requiring a similar induced decision rule between the original and 
the entire auxiliary data. Thus, our approach gives a solution to the challenging problem of enabling knowledge transfer 
from multiple {\it small} auxiliary sets with each inducing a potentially different decision rule from that on the 
original data.
\\
\\
Next, we conduct some experiments. We first produce a visualization of the original training set (corresponding to 
breast cancer) and that enhanced by transfer learning from other cancer types, including those associated with 
NMB, CK56 and CD117. Each image in the training set can be viewed as a point in the high dimensional space, and 
the points are plotted along the first and second component from a principal component analysis \cite{HTF2001} 
of the data. From Figure~\ref{figure:pcaVisual}, it can be seen that for the enhanced training set, the separation 
of points become larger. 
In particular, points with color {\it orange} now becomes visible (previously they are mostly hiding among 
points with other colors or labels); some blue points are also better separated from the green and red point clouds. A better separation
between classes would make the classification task easier, thus a better accuracy can be expected. Indeed, we can get a more
precise characterization of the amount of class separation by the class separation ratio
\begin{equation}
\rho = \sum_{\text{All pairs of classes i, j}} (SSW_i + SSW_j)/SSB_{i,j},
\end{equation}
which is the ratio of the within-class sum of squared distances out of the between-class sum of squared distances calculated over all pairs of classes, 
and $SSW_i$ and $SSB_{i,j}$ are defined as
\begin{eqnarray*}
SSW_i &=& \sum_{\text{$a, b$ all with label $i$}} distance^2(a, b)\\
SSB_{i,j} &=& \sum_{\text{$a$ with label $i$, $b$ with label $j$}} distance^2(a, b),
\end{eqnarray*}
where $distance(a,b)$ is the Euclidean distance between points $a$ and $b$. The class separation ratio $\rho$ 
measures the quality of clustering, so it gives hint on the difficulty in 
separating different classes. A smaller value of $\rho$ indicates that the within-class distances is small relative to the between-class 
distance, thus a better separation of classes. The mean class separation ratio is calculated as 15.6668 and 12.1379, averaged over 100 runs 
on the original and transfer learning enhanced training sets, respectively. This implies that, for the enlarged training set,  different classes are 
better separated. This is consistent with our visualization and experimental results, thus giving empirical 
support to the transfer learning scheme we propose. Further experiments are expected to better understand this, which we leave 
for future work.
\section{Conclusions}
\label{section:conclusions}
An algorithm has been proposed for the scoring of TMA images via transfer learning. 
By selectively including TMA images with similar staining patterns from other cancer types, 
the algorithm is able to achieve the accuracy level of a pathologist. 
This algorithm is fully automatic and without involvements from the pathologists. 
One desirable feature of the proposed algorithm is that as more transferable cancer types are considered, the accuracy is improved further. 
It is interesting to note that the accuracy would suffer if we simply combine TMA images from other cancer types without 
transfer learning. With this algorithm, pathologists are expected to diagnose cancer patients faster, more accurately, 
and consistently. Survival rates can be significantly improved because diagnoses can now be made in real-time and patients 
can be treated earlier. 
\\
\\
It is worthwhile to note that we implement transfer learning in a nonstandard setting---the auxiliary set is small and there are potentially
multiple auxiliary sets available. It is challenging as typical algorithms for transfer learning are no longer applicable here, and also we wish to 
enable knowledge transfer from as many auxiliary sets as possible. Our algorithm has sound theoretical support from recent developments
in transfer learning. It can be understood as trying to carve out a portion of the auxiliary set so that the covariate-shift model applies,
i.e., the selected subset from the auxiliary set is conformal to the original hypothesis. 
\\
\\
Empirically, experiments have also been carried out to understand the algorithm. Data visualization shows that our algorithm increases 
the class separation, and a larger class separation often makes the classification problem easier and thus improved accuracy can be 
expected. This is corroborated by the empirical class (cluster) separation ratio, which measures the cluster quality, and the enhanced 
training set leads to a better class separation. One possibility of future work is to explore how to exclude unnecessary parts (or patterns) 
in TMA images or finding the most important features about TMA images to further increase the accuracy. Additionally, different notions 
of confidence may be explored for instance transfer, for example those using the concept of conformal classification \cite{VovkShafer2008}.

\section{Acknowledgments}
I would like to thank my science research teacher, Mrs. Piscitelli, and Dr. Gordon Wu for their help and advice in the project. 
I am grateful to the editor, the associate editor, and the anonymous reviewers for their constructive comments and suggestions.

\nocite{*}

\end{document}